\documentclass{article}

\usepackage[square,sort,comma,numbers]{natbib}
\usepackage{arxiv}

\usepackage[utf8]{inputenc} 
\usepackage[T1]{fontenc}    
\usepackage{hyperref}       
\usepackage{url}            
\usepackage{booktabs}       
\usepackage{amsfonts}       
\usepackage{nicefrac}       
\usepackage{microtype}      
\usepackage{lipsum}
\usepackage{float}
\usepackage{multirow}
\usepackage{graphicx}
\usepackage{amsmath}
\usepackage{appendix}

\title{Predicting user intent from search queries using both CNNs and RNNs}

\author{
  Mihai Cristian Pîrvu \\
  MorphL \\
  \texttt{mihaicristianpirvu@gmail.com} \\
 \And
 Alexandra Anghel \\
  MorphL \\
  \texttt{alexandra@morphl.io} \\
 \And
 Ciprian Borodescu \\
  MorphL \\
  \texttt{ciprian@morphl.io} \\
 \And
 Alexandru Constantin \\
 MorphL \\
 alex@devmotion.ro \\
}

\begin{document}
\maketitle

\begin{abstract}
Predicting user behaviour on a website is a difficult task, which requires the integration of multiple sources of information, such as geo-location, user profile or web surfing history. In this paper we tackle the problem of predicting the user intent, based on the queries that were used to access a certain webpage. We make no additional assumptions, such as domain detection, device used or location, and only use the word information embedded in the given query. In order to build competitive classifiers, we label a small fraction of the EDI query intent prediction dataset \cite{edi-challenge-dataset}, which is used as ground truth. Then, using various rule-based approaches, we automatically label the rest of the dataset, train the classifiers and evaluate the quality of the automatic labeling on the ground truth dataset. We used both recurrent and convolutional networks as the models, while representing the words in the query with multiple embedding methods.
\end{abstract}

\section{Introduction}
The challenge of building intelligent internet systems requires a deep understanding of how various entities interact with each other. For example, users on a website might act differently based on how they reached the website: directly typing the URL, clicking on an ad (as well as the ad content), coming from a search engine and so forth. Therefore, the main issue lies in defining these interactions and then making educated assumptions, with various degrees of certainty. Furthermore, with the increasing ways of collecting user and website data from various sources (also known as connectors), modeling their interactions becomes a problem of sorting through various signals, which can be effectively done using deep learning models. These models thrive on data, and the data is generated directly from user actions and website statistics, which makes it very reliable. An interesting topic arises from using multiple sources of data and also multiple websites at once, compared to building classifiers only tailored for one such data source.

One fundamental problem in defining user behaviour is understanding their desires based on their intent. This can be effectively done using only the keywords entered by the user in the search engine, before reaching a website. This information is already available for any website, using tools such as Google Analytics \cite{google-analytics} making this research very desirable for anybody who wants to integrate higher level knowledge into their decision making process.

We define the problem as a coarse intent prediction, splitting the possible intents in 3 categories: informational, transactional and navigational, as described in \cite{broder2002taxonomy}. The first category is about queries where the user wishes to enrich their knowledge about some subject. In the second category lie those queries where the user wishes to purchase, sell or rent some objects or knowledge. Finally, the third category is for those queries regarding some local place or some virtual website he or she wishes to arrive at. It can be seen that these categories split the queries in 3 well defined directions. This allows further modeling of the queries, such as finer sub-intents, starting from these coarse ones, as is done in \cite{jansen2010classifying}. Another important aspect is that some queries are ambiguous by default (as is natural language in general), where the context becomes very important in order to reach a final conclusion. Since we only work at query level, without any additional information, we allow queries to be part of multiple classes, thus transforming the problem into a multi-intent classification. Further information about how the data is modeled and how the predictions are done will be discussed in later sections.

A large amount of research has been done in the domain of NLP and Information Retrieval, with a big focus on subjects like text understanding using large corpora \cite{kim2016character,zhang2015text}, Part of Speech tagging \cite{collobert2011deep,santos2014learning}, Sentiment analysis, \cite{dos2014deep, nakov2016semeval} or user action prediction \cite{liu2016repeat}.

In the domain of intent prediction, two approaches are usually taken, either a rule-based one or a statistical approach. The first one consists of applying various rules at word and query level, based on previous observations and human intuition, in order to directly predict the intent, without any statistical inference required. The advantage of this approach is that there is no doubt about the result and the predicted intent can be back-traced to basic rules. However, as websites grow bigger and the user demand increases, the rules become increasingly complex, leading to reduced debugging possibilities as well as increased difficulty in adding new rules. The second approach is using statistical approaches, specifically machine learning algorithms and models, which leverage real-world annotated data in order to make predictions on new data. With this approach, the difficulty comes in finding the best set of parameters (trainable and not), constructing good evaluation metrics, but also annotating the data in a good way, such that the models can learn a distribution that will be robust to new data. We choose the second approach, and each of these problems will be discussed in the following sections, starting with the data, and introducing a weakly-supervised automatic labeling on a large dataset in Section \ref{sec:data-processing}. Then, we define the models used in solving this task in Section \ref{sec:models-discussion}, following with a presentation of the results in Section \ref{sec:results}. Finally, we end with conclusions and future work regarding this particular task.

Similar work in detecting user intent from queries was done in various articles. In \cite{wu2010identification} they use deep language features extracted from the query, as well as search engine feedback to model the words, then train an SVM using these features on a private dataset. In \cite{hernandez2012simple}, they use similar features, but on a public dataset, testing both SVM and Naive Bayes algorithms. They observe the need for multi-label classification, talking about the ambiguities of the queries and the limitations of their single-label classification approach. In \cite{das2013determining} they first label web pages and then classify user queries in the three classes, based on the web pages classifier, and fuzzy rules based on user history and search engine history. Recent articles use more advanced machine learning algorithms, such as LSTMs and word embeddings. In \cite{kim2016intent} they enrich the off-the-shelf word embeddings using various language rules, such as pushing antonyms further away, keeping synonyms together, while maintaining the original distances to the word neighbours. Then, they train a LSTM-based neural network to predict user intent on two public datasets. Finally, in \cite{liu2016attention}, they train a LSTM network for both slot-filling and intent prediction, and show that this multi modal training improves the results of both tasks, on public datasets.

Our approach is similar to the ones described previously, using various word embeddings, such as the standard one-hot encoding, the GloVe pre-trained embedding \cite{pennington2014glove} (with 100 and 300 dimension vectors) and the FastText pre-trained embedding \cite{bojanowski2016enriching} (with 300 dimension vectors). We use a private dataset, used for the EDI challenge \cite{edi-challenge-dataset} in which we annotate a small amount of it, in order to have a correct ground-truth labeled dataset. Then, we use various automatic labeling methods, train neural network models and test on this ground truth, to validate the correctness and competitiveness of the labeling procedure. For the models, we use 3 recurrent neural network models in increasing complexity and one convolutional (non-recurrent) model using the max-pooling over time procedure described in \cite{kim2014convolutional}. We present the results of the majority of these combinations, using the ground truth labeled data for both multi-intent prediction and single-intent prediction.

\section{Data processing}
\label{sec:data-processing}

\subsection{Dataset}
\label{subsec:dataset}

As previously described, throughout this article, for the conducted experiments we have used the EDI query dataset, which was used for an intent prediction competition. The dataset is provided as a CSV with 15,263,720 entries, in various languages, which was exported using the Google Ads platform.

An example of such entry can be seen in Table \ref{table:table-1} :

\begin{table}[H]
\centering
\begin{tabular}{|c|c|c|c|c|c|}
\hline
	ID group & ID Keyword & Date & Impressions & Clicks & Keyword \\
\hline
	13621622603 & 29675024492 & 20180101 & 1 & 0 & ski boat for sale \\
\hline
\end{tabular}
\caption{Dataset entry example\label{table:table-1}}
\end{table}

We have 6 columns, the first one groups various query keywords to specific (anonymized) ads groups. Then, each keyword has an unique ID and a date when the query was exported from Google Ads. The 4th and 5th columns are the number of impressions (how many times and ad was displayed when a keyword was searched) and the number of clicks (how many times were these impressions pressed by users). Finally, the last column is the keyword itself.

Out of all these columns, for this article we only use the keywords. Since the number of keywords is very high, and the languages very diverse, we split it into two smaller datasets. The first one uses only the first 1 million entries, called \textit{EDI-Intent-1M} or \textit{1M} from now on, filtering the unusable queries only for these entries, based on some conditions that will be defined later. The second one uses all the queries that are in English, using the following tool \cite{nakatani2010langdetect}, resulting in 4,503,230 entries, called further \textit{EDI-Intent-EN} or \textit{EN}. A small summary of the split can be seen in Table \ref{table:table-2} :

\begin{table}[H]
\centering
\begin{tabular}{|c|c|c|c|}
\hline
	Dataset & Entries & Percent & Info \\
\hline
	EDI-Intent-Complete & 15,263,720 & 100\% & multi-language \\
\hline
	EDI-Intent-1M & 1,000,000 & 6.55\% & multi-language, first 1 million entries \\
\hline
	EDI-Intent-EN & 4,503,230 & 29.50\% & english only \\
\hline
\end{tabular}
\caption{Dataset split summary \label{table:table-2}}
\end{table}

Most of the experiments that will be presented later are only made on the 1M dataset, due to speed of experimentation and similarity of results. We observe that the results do not improve greatly using the larger dataset, but rather they improve by carefully designed labeling rules, by using different word embeddings, as well as model used.

\subsection{Ground truth labeling}

All the results that will be presented are done on a small subset (\textasciitilde 1K entries) of the EN dataset, which was labeled manually using 3 different annotation sources. The first two of them used a multi-label approach (allowing each query to be labeled into one, two or even three intents) and the last one annotating only the most relevant intent.

Then, using these three annotations, we create two small ground truth test sets (further called \textit{GT-2} and \textit{GT-3}), first labeling intents for each query if two annotators agree, for each of the three classes. The second test set is created by storing only the queries where all three annotators agree. One particular difference between these two sets is that the first one is multi-intent, while the second one is single-intent, because the third annotation is done using only the most relevant intent. These two test sets are made public, so that further research can be done with them. A few example queries from the two sets are presented in Table \ref{table:table-3} :

\begin{table}[H]
\centering
 \begin{tabular}{|l|l|l|l||l|l|l||l|l|l||l|l|l||l|l|l|}
    \hline
    \multirow{2}{*}{Query} &
      \multicolumn{3}{c||}{Annotator 1} &
      \multicolumn{3}{c||}{Annotator 2} &
      \multicolumn{3}{c||}{Annotator 3} &
      \multicolumn{3}{c||}{GT-2} &
      \multicolumn{3}{c|}{GT-3} \\
    & I & T & N & I & T & N & I & T & N & I & T & N & I & T & N \\
    \hline
    map of maine towns & 0 & 0 & 1 & 0 & 0 & 1 & 0 & 0 & 1 & 0 & 0 & 1 & 0 & 0 & 1 \\
    \hline
    what to do hervey bay & 1 & 0 & 1 & 1 & 0 & 0 & 0 & 0 & 1 & 0.5 & 0 & 0.5 & \multicolumn{3}{c|}{n/a} \\
    \hline
    when is the best time to fish & 1 & 0 & 0 & 1 & 0 & 0 & 1 & 0 & 0 & 1 & 0 & 0 & 1 & 0 & 0 \\
    \hline
    ex demo cars for sale & 0 & 1 & 0 & 0 & 1 & 0 & 0 & 1 & 0 & 0 & 1 & 0 & 0 & 1 & 0 \\
    \hline
    new homes for sale bournemouth & 0 & 1 & 1 & 0 & 1 & 1 & 0 & 1 & 0 & 0 & 0.5 & 0.5 & \multicolumn{3}{c|}{n/a} \\
    \hline
    australia inheritance tax & 1 & 0 & 0 & 1 & 1 & 0 & 1 & 0 & 0 & 1 & 0 & 0 & \multicolumn{3}{c|}{n/a} \\
    \hline 
   banking for you & 0 & 1 & 0 & 0 & 0 & 1 & 1 & 0 & 0 & \multicolumn{3}{c||}{n/a} & \multicolumn{3}{c|}{n/a}  \\
    \hline
  \end{tabular}
\caption{Ground truth labeling examples\label{table:table-3}}
\end{table}

The first two annotators used the multi-intent labeling, while the third used single-intent labeling. It can be seen that some intents may match multiple agreements, but not match all three of them, in which case that item is removed from the labeling process. In the last example we can see a case where all the three annotators give different labels, in which case the query is removed from both sets. This kind of annotation can be seen as a sort of voting between multiple annotation sources, where only collaborative work between 2 or 3 sources can be marked as reliable. This can be seen as a method of reducing bias in the ground-truth set. A short summary of the two datasets is presented in Table \ref{table:table-4} :

\begin{table}[H]
\centering
\begin{tabular}{|c|c|c|c|c|c|}
\hline
	\multirow{2}{*}{Dataset} & \multirow{2}{*}{Entries} & \multirow{2}{*}{Info} & \multicolumn{3}{c|}{Counts} \\
	& & & Informational & Transactional & Navigational \\
\hline
	GT-2 & 1,197 & multi-intent, 2 agreements & 383 & 502 & 474 \\
\hline
	GT-3 & 696 & single-intent, 3 agreements & 200 & 265 & 231 \\
\hline
\end{tabular}
\caption{Ground-truth test sets summary. Note that the multi-intent dataset (GT-2) counts each intent independently, thus the sum being larger than the number of entries. For the single-intent test set (GT-3), the sum of the counts is equal to the number of entries. \label{table:table-4}}
\end{table}

The results that will be presented will include numbers for both ground-truth sets, in order to provide insights about both single-intent and multi-intent prediction capabilities of the model.

\subsection{Automatic labeling}

In order to train the models used throughout this paper, we require a labeled set, which can be used as training data, in order to predict the intent from the queries alone. Since we do not have any annotated data, we resorted to a two-step process. The first step is consisted in an automatic labeling method, following various rule-based methods. For this paper, we only used word-level rules, without incorporating any query (utterance) information. The rules, which will be explained later, were simply checking various keywords, or looking for similar words to those keywords, which we manually added to the set of belonging to one of the three possible intents. For example, queries containing words like \textit{buy} or \textit{rent} will be labeled as transactional, because these keywords belong in the transactional set. We then apply these rules to both 1M and EN datasets, removing all the entries that do not contain any words in the keyword sets, or words that do not have embeddings according to the chosen embedding dictionary (more about this in the next section). Using this labeled dataset (be it 1M or EN), we train various recurrent and non-recurrent models, pick the best model checkpoint based on the validation set, and then test it on both ground truth sets (GT-2 and GT-3). Based on the score reached by each model, we validate the quality of the annotation, as well as the complexity of the model or word embedding choice.

While it may seem very naive to use this word level annotation rules, we observe that the quality of the results improves drastically by simply constructing well-thought keyword sets, compared to using other rule-based approaches from the literature.

We have constructed multiple rule-based methods, 8 in total, starting from a very basic set of words, and then improving each of this method iteratively, based both on results and common sense. We will call each of this method $V_{i}$, where $i \in {1..8}$. We also used an external annotation tool \cite{twinword-annotation-tool}, which we will call further Ext-1 (external annotation method 1). In what follows, we will described succinctly each of the 8 proposed methods and then provide a compiled table which highlight the similarities and differences between each of them. 

The initial method (V1) was derived using the rules described in \cite{jansen2010classifying}, with a single-intent labeling (based on the last word in the query, which is present in the keyword sets). Then, starting from this approach, the second version (V2) simply allowed multi-intent labeling. Basically, if a query had multiple keywords in different sets, then each of them contributed to the labeling of the query (not just the last one, as in V1). We observed that by just this simple change, a small improvement is obtained (which makes sense, because GT-2 is also multi-intent). Since this approach simply consists of taking various keywords from a different source, we were limited to their vocabulary, which meant dropping a large amount of queries from our data, limiting the effective training sets.

Thus, the third version (V3) tries to combine the two approaches we used, combining the words from V1 \& V2 with the words labeled using Ext-1. In order to acquire the necessary data, we first label the 1M dataset using Ext-1, then perform statistics on the labeled words. We take the top 50 words in each of the 3 intents, based on the automatic labeling, and add those to our keyword sets, thus increasing the dictionary we use to label the data. We also ignore the english stop words, since these bring close to no information about the intent of the query, using the list from \cite{bird2009natural}. However, we observe that this method performs poorer than the more simplistic version, mostly because the words from the gathered statistic are overlapping in multiple intents. A side effect of this is that the models learn to over-predict multi-label intents for almost any query, thus lowering the confidence in single-intents (which are more dominant in number), increasing the loss score and lowering the accuracy on the ground truth set.

At this moment, we decided to use a more common sense approach, manually adding and removing words from V1. This proved to be the most effective improvement overall, as we'll present in the Results section. A complete list of the words used for the V4 automatic labeling process can be seen in Table \ref{table:table-5} :

\newpage

\begin{table}[H]
\centering
\begin{tabular}{|c|l|}
\hline
	Intent & Keywords \\
\hline
	Informational & what is, how, how much, how many, vs, when, testimony, testimonial, testimonial, testimonies, list, \\
& compare, comparison, playlist, playlists, review, reviews, types, diet, diets, beauty, recipe, recipes, \\
& tip, tips, trick, tricks, exercise, exercises, technique, techniques, diy, best, lyric, lyrics, horoscope, \\
& horoscopes, craft, crafts, joke, jokes, story, stories, humor, walkthrough, graph, graphs, article, \\
& articles, party, definition, cause, causes, new, shares, tax, worth, grow, plant, write, cook, study, \\
& information, book, books, top, many, idea, ideas, meaning, mean, tool, tools, art, care, business, land, \\
& music, letter, calorie, calories, ounce, pound, pounds, kg, kilo, kilos, question, questions, spoon, \\
& spoons, gram, grams, ton, tons, yard, yards, feet, metre, metres, inch, inches, pic, pics, picutre, \\
& pictures, image, images, gallery, galleries, menu, mortgage, income, detox, plan \\
\hline
	Transactional & weather, forecast, centre, st, bus, route, routes, train, station, shop, gps, location, job, jobs, store, \\
& stores, far, market, supermarket, land, bank, university, gov, company, com, co, mil, corporation, \\
& www, association, dealer, zip, zip code, area, cruise, in, street, net, society, org, inc, mi, academy, edu, \\
& http, .au, .ca, .de, .eu, .fr, .jp, .us, .uk, where, close, near, nearby,  map, maps, time \\
\hline
	Navigational & watch, tv, read, buy, sell, sale, price, prices, pay, paid, money, cost, free, cheap, order, much, \\
& used, purchase, model, models, clean, remove, quick, vacuum, build, share, use, write, get, make, \\ 
& download, downloads, rar, chat, software, softwares, convert, wallpaper, wallpapers, chart, charts, \\
& game, games, application, applications, app, apps, course, courses, repair, repais, shop, rent, rentals, \\
& job, jobs, digital, number, control, best, store, stores, care, credit card, check, rate, rates, online, \\
& online shopping, visa, gift, gifts, car, cars, in bulk, subscription, subscriptions, free shipping, \\
& shipping, change, coupon, coupons, ticket, tickets, financing, interest, dealer, top up, pay as you go, \\
& parts, part, cruise, shopping, mortgage, converter, convertor, clothes \\
\hline
\end{tabular}
\caption{Word-based rules for V4 auto-labeling process\label{table:table-5}}
\end{table}

We can observe that the number of keywords is roughly similar in all three categories, with words being chosen by both common sense and experience using the 1M and EN datasets.

Moving forward from this, we incorporated synonyms for V4 keywords, in the V5 algorithm. This only provided marginal variations in results, from which we can understand that synonyms, while providing the same meaning at word level, can influence the intent.

The V6 version used the words from V5 (thus, including synonyms), but instead of looking for specific word matches, we used word embeddings and computed distances between each keyword and each word in the query. If this distance was below a threshold, we concluded that the words were very similar (contextual similarities), and we included those matches as they would be part of that particular set. The used embedding was GloVe (trained on Wikipedia 2014 + Gigaword 5, called 6B, 100 dimensions), the distance was computed using the squared L2: $ distance = \sum_{i=1}^{100} (E(w_{1})_{i} - E(w_{2})_{i})^{2} $ and the threshold was empirically chosen at 20. This version also provided a small variation in the results, compared to V4.

Going forward, the V7 version, improves on V6 by prioritizing exact matches and removing test set words. Basically, if words are exact matches (distance is 0), then this is identical to V5. However, if no words are identical, the choice is made like it would be V6. The embedding and threshold were also kept identical. This version proved much better, improving the intent prediction with about 3-4\% compared to the V4 version (which is the 2nd performing algorithm).

The last version, V8, tries to improve on V7, by variating the embedding, threshold and distance used, by the means of grid searching. The metric, however, is not training various models and comparing the result on ground truth, but rather comparing the Hamming Distance directly on the ground truth set (GT-2). Unfortunately, this method performed much poorer than the other versions, which means that further analysis must be done, such as checking each combination by training $ E \times T \times D $ models and then testing on GT-2 and GT-3. E can be any embedding (GloVe 6B 100, GloVe 6B 300, Fasttext, One-hot etc.), thresholds can vary in an infinite modes (so picking good guesses is a requirement here) and competent distances must also be chosen well (L2, L1, cosine similarity etc.).

A summary of all the automatic labeling methods can be seen in Table \ref{table:table-6} :

\newpage

\begin{table}[H]
\centering
\begin{tabular}{|c|c|c|c|c|c|c|}
\hline
	\multirow{2}{*}{Labeling} & \multirow{2}{*}{Info} & \multirow{2}{*}{Multi intent} & \multirow{2}{*}{1M count entries} & \multicolumn{3}{c|}{Counts} \\
	& & & & Informational & Transactional & Navigational \\
\hline
	V1 & rules from \cite{jansen2010classifying} & No & 101,216 & 53,865 & 42,158 & 5,193 \\
\hline
	V2 & rules from \cite{jansen2010classifying} & Yes & 89,698 & 49,395 & 38,406 & 4,739 \\
\hline
	V3 & rules from \cite{jansen2010classifying} and statistics & Yes & 224,858 & 109,219 & 148,154 & 120,202 \\
	& from Ext-1 (top 50 words) & & & & & \\
\hline
	\textbf{V4} & keywords from Table \ref{table:table-5} & Yes & 197,837 & 56,812 & 111,500 & 70,738 \\
\hline
	V5 & V4 + synonyms & Yes & 23,3217 & 87,672 & 126,248 & 76,227 \\
\hline
	V6 & V5 + GloVe embeddings & Yes & 361,162 & 179,326 & 181,095 & 105,959 \\
\hline
	\textbf{V7} & V5 + GloVe embeddings & Yes & 389,775 & 178,748 & 177,152 & 87,776 \\
	& + prioritize exact matches & & & & & \\
\hline
	V8 & grid search for embedding, & Yes & 358,021 & 162,409 & 212,940 & 115,461 \\
	& distance and threshold & & & & & \\
\hline
	Ext-1 \cite{twinword-annotation-tool} & external annotation tool & Yes & 34,252 & 2,045 & 8,916 & 27,796 \\
\hline
\end{tabular}
\caption{Statistics of various automatic labeling algorithms, on the 1M dataset. For V6, V7, GloVe-6B-100 embedding was used, while for V8 GloVe-6B-300 was used using cosine similarity and a threshold of 0.56. \label{table:table-6}}
\end{table}

In our findings, we observed that only Ext-1, V4 and V7 automatic labeling provided competitive predictions, thus the majority of the results that will be provided will only include values for these annotation rules. We can observe that for Ext-1 the results are heavily biased towards navigational, but still provide accurate predictions, which proves that a good annotation is much better than a mediocre one, but with lots of data.

\section{Models discussion}
\label{sec:models-discussion}

This section focuses on the architectures of the models used for training the data, which was previously discussed. When working with utterances or queries as in this case, it is very natural to structure the model in a recurrent way, because of the temporal order of the words, which may influence the results, rather than treating them independently. Thus, a linear recurrent neural network model was the first thing we used to model the intent prediction problem. Then, we increased the complexity of the model, in order to see if adding non-linearities, more complicated recurrent layers (LSTM) or stacking multiple recurrences would help this problem. In the end we have 3 recurrent models, each in increasing complexity compared to the previous one. On the other end, we also built a convolutional neural network (with no recursion), where each word in the query is independent from the previous ones, using max-pooling over time, inspired from \cite{kim2014convolutional}. Conceptually, each word is filtered using a convolutional layer and then for each index in the feature of the words, we compute the maximum value, in the end resulting in a single feature over the entire query.

The input for the neural networks is a word embedding (GloVe, Fasttext, One-hot) and they produce a probabilistic output for each of the three intents, from a softmax activation function. Then, during test time, the chosen intent is the one with the highest probability (single-intent prediction), or the probabilities themselves (multi-intent prediction).

\subsection{Recurrent Models}

We'll first describe the three recurrent models, which will be further called RNN-i, $i \in {1, 2, 3}$. For each such model, we'll present a top level figure, highlighting the main differences between each of them. Then, we'll discuss the chosen sizes for the layers and present the number of parameters based on various embedding choices. Since all three models are incremental changes to the basic one, there is a large overlap between them. The first one is completely linear, the second one adds non-linearities and another FC layer and the third one adds LSTM layers and two recurrent layers, as well as a third FC layer with non-linearities.

\newpage

\subsubsection{RNN-1}
The architecture of the first model can be seen in Figure \ref{fig:fig-1} :

\begin{figure}[H]
\centering
\includegraphics[scale=0.25]{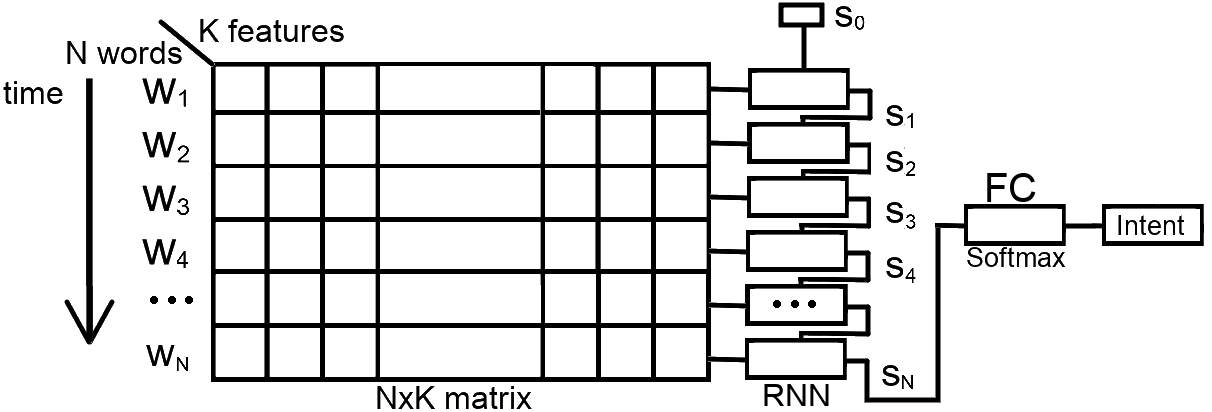}
\caption{\label{fig:fig-1} Architecture of RNN-1 model.}
\end{figure}

As described earlier, each word feature (denoted as a K-feature embedding here) is concatenated with the hidden state of the recurrent model. In the figure, the recurrent model is shown unfolded in time, but it should be noted that the matrices are shared and weights are updated using the standard BPTT algorithm. Then, the last hidden state ($s_{N}$) which corresponds to the output of the Nth word in the query and the N-1th state of the network is fed into a fully-connected layer (with no activation function), which outputs three values, corresponding to the three intents. This values are passed through a softmax function, in order to normalize them to probabilities and enable efficient learning. The number of dimensions of the words embedding can vary (100 for GloVe-100, 300 for GloVe-100 or Fasttext and 48,692 for one-hot). The hidden state size was chosen as to having 101 dimensions. Thus, the last FC layer has a 101x3 matrix size (+ biases).

In Table \ref{table:table-7} we present the number of trainable parameters based on the type of embedding:

\begin{table}[H]
\centering
\begin{tabular}{|c|c|c|}
\hline
	Embedding type & Feature size & Parameters \\
\hline
	GloVe-100 & 100 & 20,809 \\
\hline
	GloVe-300 \& Fasttext & 300 & 41,009 \\
\hline
	One-Hot & 48,692 & 4,928,601 \\
\hline
\end{tabular}
\caption{RNN-1 parameters count based on type of embedding \label{table:table-7}}
\end{table}

We can see that, when using a pre-trained embedding, the network must learn much fewer parameters, compared to the case where each word is represented as a one-hot encoding. However, as we'll see, using this naive representation gives the best results.

\subsubsection{RNN-2}

The second model is a natural evolution of the first, where we use a ReLU non-linearity from the last hidden state to the first FC layer. We also include a secondary FC layer, which is supposed to project the query into a domain that can capture higher level information. This idea is what made neural network models popular, but as we'll see, the results don't change that much compared to RNN-1 model. The last FC outputs again 3 numbers, which are passed through a softmax activation function that outputs a probability map for the three possible intents. The size of the hidden state vector was empirically chosen at 100.

\begin{figure}[H]
\centering
\includegraphics[scale=0.25]{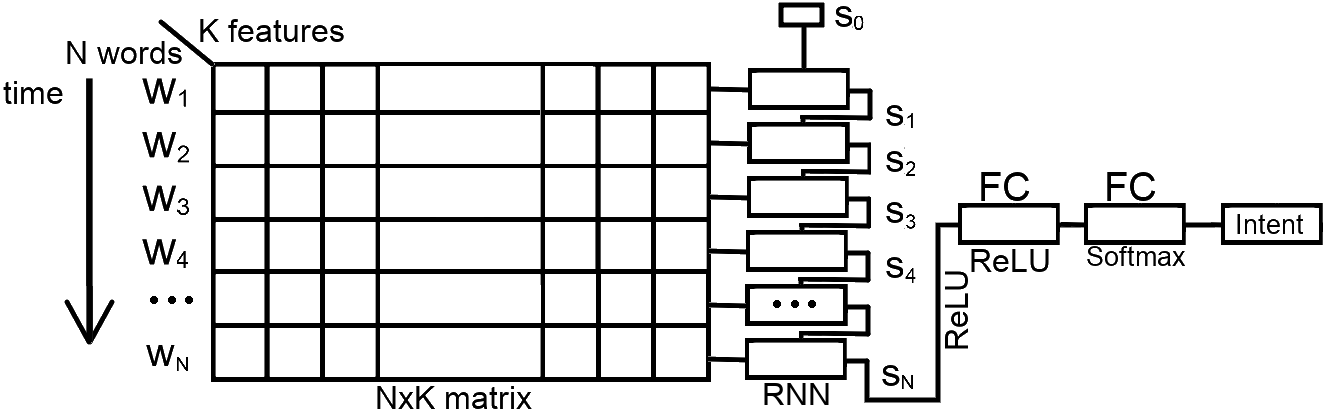}
\caption{\label{fig:fig-2} Architecture of RNN-2 model.}
\end{figure}

In Table \ref{table:table-8} the number of trainable parameters for RNN-2 is presented:

\begin{table}[H]
\centering
\begin{tabular}{|c|c|c|}
\hline
	Embedding type & Feature size & Parameters \\
\hline
	GloVe-100 & 100 & 20,809 \\
\hline
	GloVe-300 \& Fasttext & 300 & 41,009 \\
\hline
	One-Hot & 48,692 & 4,928,601 \\
\hline
\end{tabular}
\caption{RNN-2 parameters count based on type of embedding \label{table:table-8}}
\end{table}

\subsubsection{RNN-3}

The third and last recurrent model adds two changes to the basic architecture, that are supposed to capture higher level information about the queries. First, we use a LSTM model \cite{hochreiter1997long}, as opposed to a simple RNN. Secondly, we stack two such LSTMs sequentially, which means that each hidden state vector at any timestamp is fed into a secondary LSTM. The last state of this second LSTM is then fed into a fully-connected layer, after being passed through a ReLU activation. We also add a third FC layer, and similarly to the previous models, output three vectors, that are transformed to probabilities using a softmax function.

\begin{figure}[H]
\centering
\includegraphics[scale=0.25]{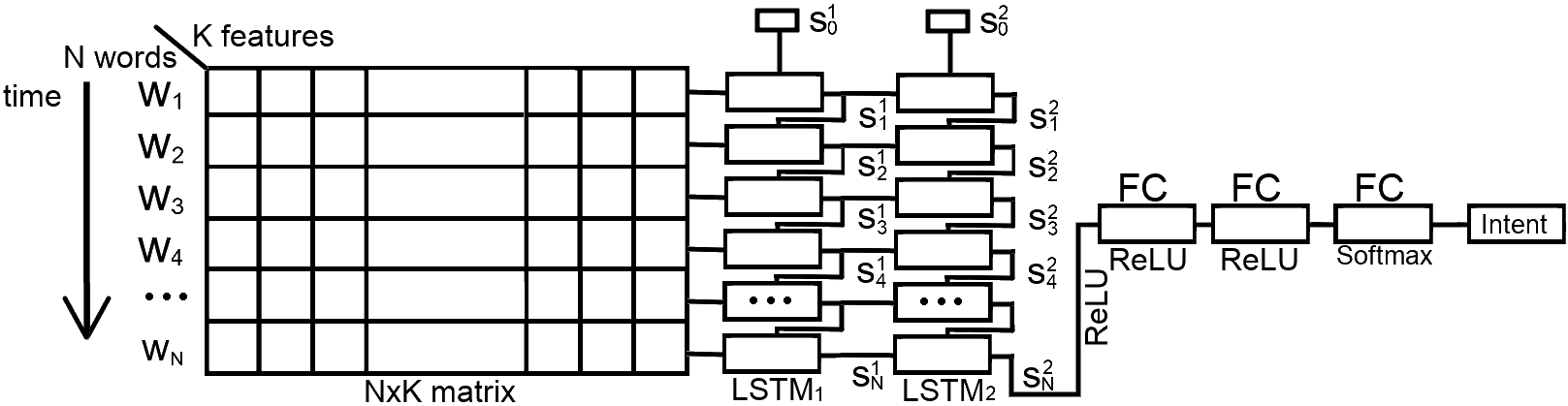}
\caption{\label{fig:fig-3} Architecture of RNN-3 model.}
\end{figure}

In Table \ref{table:table-9} the number of trainable parameters for RNN-3 is presented:

\begin{table}[H]
\centering
\begin{tabular}{|c|c|c|}
\hline
	Embedding type & Feature size & Parameters \\
\hline
	GloVe-100 & 100 & 182,103 \\
\hline
	GloVe-300 \& Fasttext & 300 & 262,103 \\
\hline
	One-Hot & 48,692 & 19,618,903 \\
\hline
\end{tabular}
\caption{RNN-3 parameters count based on type of embedding \label{table:table-9}}
\end{table}

We can see that the number of parameters increases for each model and for each embedding. For this particular model, we only trained using the GloVe-100 embedding as we have seen that the improvement is non-existent when increasing the model complexity to this problem.

\subsection{Convolutional Model}

The last and final model we used is a non-recurrent model, which uses 2D convolutions in order to filter multiple words at once. Multiple independent convolutions were used having different shapes, which has the effect of filtering multiples words at the same time. For the embedding depth, a constant of 3 was kept, so all the convolutions had a shape of $i \times 3, i \in {1...n}$, where i represents the number of words filtered at the same time. Conceptually, this tries to combine features of multiple words, in order to increase the window of the query to capture longer context. All these features were concatenated together, resulting in a large matrix, with feature indexes corresponding to one word and a various context sizes, from 1 to n. The next step was keeping just the most relevant feature index of each feature map, which is equivalent to only keeping the most relevant features of the most relevant words. This operation was done using max-pooling over time (in our case max-pooling over words), inspired by \cite{kim2014convolutional}. The resulting feature map was then passed through a 1D convolution, in order to decrease the size and then passed through a fully-connected layer. This fully-connected layer produced three numbers, similarly to the recurrent models, which were finally normalized to probabilities using a softmax activation. Each convolution was followed by a ReLU activation, making the model non-linear. The architecture of this model can be seen in Figure \ref{fig:fig-4} :

\begin{figure}[H]
\centering
\includegraphics[scale=0.3]{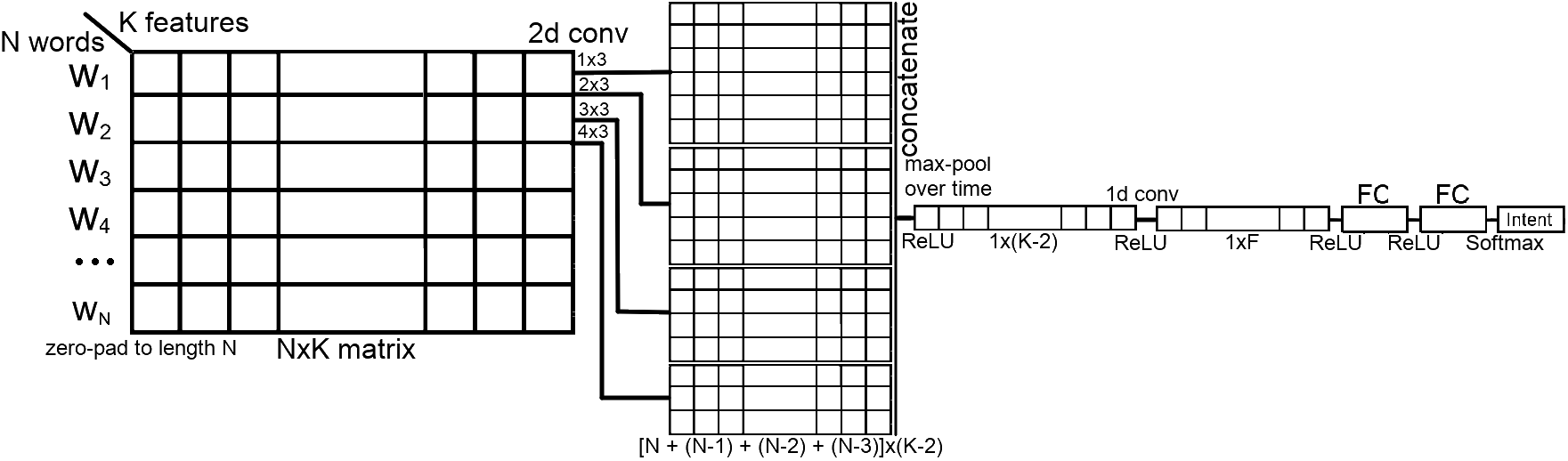}
\caption{\label{fig:fig-4} Architecture of CNN-1 model.}
\end{figure}

For the experiments ran, we used 4 convolution operations with kernel filters of shape 1x3, 2x3, 3x3 and 4x3, thus filtering 1 word, 2 words, 3 words and 4 words simultaneously. The number of feature maps was set to 100 for each operation, which were concatenated on the time (word features) axis. The sequence size was set at 4 as well, meaning that only queries of 1, 2 or 3 words were padded with zeros. The filter of the 1D convolution had a size of 1, and filtered the 100 feature maps to 10 feature maps. The first FC layer procuded a 50 dimension feature map, which is then passed through a final FC layer that produces the three intent numbers.

Finally, we present the number of parameters for this model using various embeddings in Table \ref{table:table-10} :

\begin{table}[H]
\centering
\begin{tabular}{|c|c|c|}
\hline
	Embedding type & Feature size & Parameters \\
\hline
	GloVe-100 & 100 & 53,613 \\
\hline
	GloVe-300 \& Fasttext & 300 & 153,613 \\
\hline
	One-Hot & 48,692 & 24,349,613 \\
\hline
\end{tabular}
\caption{CNN-1 parameters count based on type of embedding \label{table:table-10}}
\end{table}

We can see that the number of parameters is higher than the RNN variants, but since the model is not recurrent it is much more stable during training. In the results section, which follows, we'll see that even this simplistic model offers decent results, and even better in some cases compared to the recurrent counterparts.

\section{Results}
\label{sec:results}

This section will first talk about the setup that was used in order to produce the results. Then, we'll talk about how the results will be presented, which datasets, models and embeddings were used and how they influenced the final results. Finally, we'll provide a few qualitative results, as well as a few failing cases.

The models were built using the PyTorch deep learning framework \cite{paszke2017automatic}. The networks were trained for 20 epochs, using SGD with a learning rate of 0.01 and a momentum of 0.9. The datasets (1M and EN) were split into a 80\%-20\% train/validation split. The results that are shown later are the ones from the best performing epoch (out of the 20 epochs) on the validation set. The models were optimized using the cross-enetropy loss function, which can be expressed as:

$$ L(y, t) = -\sum_{c=\{I,T,N\}}{t_{i, j}^{(c)} * log(y_{i, j}^{(c)})} $$

Here, y is the prediction vector, t is the target vector and c is the channels for each of the three intents: informational, transactional and navigational. The vectors are normalized from a softmax layer, thus summing to 1.


The main metric that was used is a multi-modal accuracy, which can be described as:

\begin{minipage}{0.45\textwidth}
\[
Acc(y, t) =
	\left\{
        \begin{array}{ll}
			1, if \ y_{c} > thr \ and \ t_{c} > 0, \ \forall c \in \{I,T,N\} \\
			0, otherwise \\
		\end{array}
	\right.
\]
\end{minipage}%
\begin{minipage}{0.45\textwidth}
\[
thr =
	\left\{
        \begin{array}{ll}
			0.5, if \ 1 \ intent \\
			0.25, if \ 2 \ intents \\
			0.16, if \ 3 \ intents \\
		\end{array}
	\right.
\]
\end{minipage}%

Basically, this accuracy can be seen as an extension of the case with just one prediction, where the threshold is set at 0.5. Now, the threshold varies for multiple predictions, and we only care if any of these predictions are right.

The first set of experiments that are presented will offer insight about the quality of the annotation, to find out which of the methods for automatic labeling performed the best. This will be done by keeping the embedding fixed (GloVe 6B 100) as well as the model (RNN-1) and testing only on 1M dataset. Then, we'll try to use only the best performing ones, and vary the other parameters. These results can be seen in Table \ref{table:table-11} :

\begin{table}[H]
\centering
\begin{tabular}{|c|c|c|}
\hline
	\multirow{2}{*}{Labeling} & \multicolumn{2}{c|}{Accuracy} \\
	& GT-2 & GT-3 \\
\hline
	V1 & 54.64\% & 59.00\% \\
\hline
	V2 & 55.89\% & 60.01\% \\
\hline
	V3 & 50.23\% & 47.15\% \\
\hline
	V4 & 65.41\% & 69.58\% \\
\hline
	V5 & 64.63\% & 67.92\% \\
\hline
	V6 & 63.15\% & 65.42\% \\
\hline
	V7 & \textbf{67.19\%} & 70.47\% \\
\hline
	V8 & 56.38\% & 55.28\% \\
\hline
	Ext-1 & 66.69\% & \textbf{71.63\%} \\
\hline
\end{tabular}
\caption{Results for dataset 1M, on RNN-1 with GloVe 6B 100 embedding. Ext-1 results are show for comparison. \label{table:table-11}}
\end{table}

We can see clearly how much V4 and onward improves compared to the simple rules of \cite{jansen2010classifying}. We also see how important a good annotation is with the result of Ext-1, which uses just a small fraction of the data, as we can see in Table \ref{table:table-6}. Moving on, we'll just use V4, V7 and Ext-1 for annotating.

The results for the EN dataset, for the two labeling processes can be seen in Table \ref{table:table-10} :

\begin{table}[H]
\centering
\begin{tabular}{|c|c|c|}
\hline
	\multirow{2}{*}{Labeling} & \multicolumn{2}{c|}{Accuracy} \\
	& GT-2 & GT-3 \\
\hline
	V4 & 62.95\% & 66.07\% \\
\hline
	V7 & 68.02\% & 70.03\% \\
\hline
\end{tabular}
\caption{Results for dataset EN, on RNN-1 with GloVe 6B 100 embedding. \label{table:table-12}}
\end{table}

We see that the results are every similar, with V4 being a few percents below 1M, but V7 being almost one percent up. This means, however, that a partial annotation does not provide good results even when using much more data. For this reason alone, most of the results that will be presented will only be models trained on the 1M dataset or Ext-1.

Next, in Table \ref{table:table-11} we'll see how varying the model, for a particular embedding and dataset affects the results:

\begin{table}[H]
\centering
\begin{tabular}{|c|c|c|c|c|}
\hline
	\multirow{2}{*}{Model} & \multirow{2}{*}{Labeling} & \multirow{2}{*}{Dataset} &  \multicolumn{2}{c|}{Accuracy} \\
	& & & GT-2 & GT-3 \\
\hline
	RNN-1 & V4 & 1M & 65.41\% & 69.58\% \\
\hline
	RNN-2 & V4 & 1M & 66.63\% & 68.07\% \\
\hline
	RNN-1 & V7 & 1M & 67.19\% & 70.47\% \\
\hline
	RNN-2 & V7 & 1M & 66.73\% & 69.92\% \\
\hline
	CNN-1 & V7 & 1M & 66.20\% & 70.17\% \\
\hline
	RNN-1 & V4 & EN & 62.95\% & 66.07\% \\
\hline
	RNN-2 & V4 & EN & 65.05\% & 65.87\% \\
\hline
	RNN-1 & V7 & EN & 68.02\% & 70.03\% \\
\hline
	RNN-2 & V7 & EN & \textbf{69.14\%} & \textbf{72.47\%} \\
\hline
	RNN-3 & V7 & EN & 67.61\% & 70.02\% \\
\hline
	RNN-1 & \multicolumn{2}{c|}{Ext-1} & 66.69\% & 71.63\% \\
\hline
	RNN-2 & \multicolumn{2}{c|}{Ext-1} & 66.70\% & 71.14\% \\
\hline
	CNN-1 & \multicolumn{2}{c|}{Ext-1} & 67.39\% & 71.88\% \\
\hline
\end{tabular}
\caption{Results for datasets EN, 1M and Ext-1, all models and GloVe 6B 100 embedding. \label{table:table-13}}
\end{table}

We see that the bigger dataset and the more complex model does indeed improve the results slightly. However, all of them are competitive enough to make the assumption that model complexity is not the most relevant factor for this task. For example, all models for Ext-1 labeling are within 1\% accuracy for each model.

Finally, we'll see how varying the embedding, on a fixed dataset (1M) affects the results. For this set of experiments we'll test 4 potential embeddings: GloVe 6B (100 and 300 dims), Fasttext (300 dims) and One-hot encoding. The main advantage of using pre-trained embeddings is that the words are compressed into a fixed high dimension vector, independent of the number of words in the dictionary. Also, two words in such embedding space are closer in distance if they are similar than two unrelated words as well as one can perform simple arithmetic on word embeddings for word associations. These features have boosted a lot of tasks in the NLP domain, but as we will see for the intent classification task, using one-hot encoding (independence of words, and higher dimension based on the vocabulary size), the results are increased even further.

These results can be seen in Table \ref{table:table-14} :

\begin{table}[H]
\centering
\makebox[\textwidth]{\makebox[1.25\textwidth]{%
\begin{minipage}{0.51\textwidth}
\begin{tabular}{|c|c|c|c|c|}
\hline
	\multirow{2}{*}{Model} & \multirow{2}{*}{Labeling} & \multirow{2}{*}{Embedding} &  \multicolumn{2}{c|}{Accuracy} \\
	& & & GT-2 & GT-3 \\
\hline
	RNN-1 & V4 & GloVe-100 & 65.41\% & 69.58\% \\
\hline
	RNN-1 & V4 & GloVe-300 & 68.76\% & 71.69\% \\
\hline
	RNN-1 & V4 & Fasttext & 66.13\% & 66.11\% \\
\hline
	RNN-1 & V4 & One-hot & 63.44\% & 57.79\% \\
\hline
	RNN-1 & V7 & GloVe-100 & 67.19\% & 70.47\% \\
\hline
	RNN-1 & V7 & GloVe-300 & 68.55\% & 71.88\% \\
\hline
	RNN-1 & V7 & Fasttext & 69.47\% & 73.65\% \\
\hline
	RNN-1 & V7 & One-hot & \textbf{75.61\%} & 77.09\% \\
\hline
	RNN-2 & V7 & GloVe-100 & 66.73\% & 69.92\% \\
\hline
	RNN-2 & V7 & One-hot & 74.07\% & \textbf{79.01\%} \\
\hline
	CNN-1 & V7 & GloVe-100 & 66.20\% & 70.17\% \\
\hline
\end{tabular}
\end{minipage}

\begin{minipage}{0.47\textwidth}
\begin{tabular}{|c|c|c|c|c|}
\hline
	\multirow{2}{*}{Model} & \multirow{2}{*}{Labeling} & \multirow{2}{*}{Embedding} &  \multicolumn{2}{c|}{Accuracy} \\
	& & & GT-2 & GT-3 \\
\hline
	CNN-1 & V7 & GloVe-300 & 67.21\% & 69.96\% \\
\hline
	CNN-1 & V7 & Fasttext & 64.08\% & 66.42\% \\
\hline
	CNN-1 & V7 & One-hot & n/a & n/a \\
\hline
	RNN-1 & Ext-1 & GloVe-100 & 66.69\% & 71.63\% \\
\hline
	RNN-1 & Ext-1 & GloVe-300 & 68.35\% & 72.68\% \\
\hline
	RNN-1 & Ext-1 & Fasttext & 70.94\% & 72.12\% \\
\hline
	RNN-1 & Ext-1 & One-hot & 70.30\% & 74.48\% \\
\hline
	CNN-1 & Ext-1 & GloVe-100 & 66.20\% & 70.17\% \\
\hline
	CNN-1 & Ext-1 & GloVe-300 & 67.21\% & 69.96\% \\
\hline
	CNN-1 & Ext-1 & Fasttext & 64.08\% & 66.42\% \\
\hline
	CNN-1 & Ext-1 & One-hot & n/a & n/a \\
\hline
\end{tabular}
\end{minipage}}}

\caption{Results for datasets 1M and Ext-1, varying word embedding. \label{table:table-14}}
\end{table}

We can now see that the embedding does indeed influence the final result the most. While varying the model brings a very small improvement, using a richer word representation improves the results by a few percents every time. Looking at GloVe-100 vs GloVe-300, we see a constant improvement. Using Fasttext or GloVe doesn't influence that much the final result, both yielding competitive results. However, perhaps the most surprising result, is that using a non-pretrained embedding (one-hot) results in a very big improvement compared to all the other representations. However, only the recurrent models could be trained using this representation, while the convolutional model diverges during training. This may be because of the large amount of parameters required, bad initialization or lack of proper hyperparameter tuning. Either way, both RNN-1 and RNN-2 trained with one-hot encoding on the 1M dictionary yield very good results: 75.61\% for multi-intent (2 agreements) and 79.01\% for single-intent (3 agreements).

In ending, we present the results of using the automatically labeling methods on the test set, in order to show the performance of the learning methods compared to simple rule-based solutions. It should be noted that not all items in the test set can be labeled, so we also include the percentage of items that could be labeled. The GT-2 set has 1197 entries, while GT-3 has 696.

\begin{table}[H]
\centering
\begin{tabular}{|c|c|c|c|c|}
\hline
	\multirow{2}{*}{Labeling} & \multicolumn{2}{c|}{Accuracy} & \multicolumn{2}{c|}{Percentage labeled} \\
	& GT-2 & GT-3 & GT-2 & GT-3 \\
\hline
	V1 & 45.50\% & 48.66\% & 33.41\% & 43.10\% \\
\hline
	V2 & 46.25\% & 45.33\% & 33.41\% & 43.10\% \\
\hline
	V3 & 22.64\% & 7.50\% & 77.10\% & 84.19\% \\
\hline
	V4 & 32.07\% & 26.37\% & 70.59\% & 78.44\% \\
\hline
	V5 & 30.48\% & 24.74\% & 77.27\% & 84.77\% \\
\hline
	V6 & \textbf{67.83\%} & \textbf{70.78\%} & \textbf{80.78\%} & \textbf{86.06\%} \\
\hline
	V7 & \textbf{67.83\%} & \textbf{70.78\%} & \textbf{80.78\%} & \textbf{86.06\%} \\
\hline
	V8 & 51.62\% & 43.07\% & 51.46\% & 56.03\% \\
\hline
\end{tabular}
\caption{Results for GT-2 and GT-3 using V1-V8 labeling. V8 used GloVe-100, cosine similarity and a threshold of 0.36.}
\end{table}

We can conclude that the learning models outperform the rule-based system on both accuracy and percentage of items labeled. Of course, the rules can be continuously improved, but we have proved that using the same rules, the machine learning method generalizes better for new items. Adding better rules would only increase the accuracy of the models as well.

\section{Conclusions and Future Work}

In this article we introduced a two-step process for automatically labeling a new dataset, starting from a small ground truth, using various heuristic rules for the rest of the data, training various models and then testing the quality of the labeling on the small set. We have used a new dataset, which was introduced as part of a challenge, so no prior work was done in this direction, however we used a very well known 3-intent taxonomy from the literature. We have shown that the task of multi-intent prediction from queries can be modeled with both recurrent and convolutional neural networks, trained as a regular classification problem.

For future work, we will try to improve the labeling rules, looking further from just words alone, but treating the entire query. We will also look into using various other features, such as parts of speech, chunking, domain classification. Another interesting idea is to retrain the word embeddings on the large EN dataset, and then use these embeddings with contexts formed by using similar words, but the other columns as well (such as ID group, impressions or clicks).

In the \hyperref[sec:appendix]{Appendix}, we provide a few quantitative results from the best performing models on GT-2.

\newpage


\begin{appendices}
\section*{Appendix}
\label{sec:appendix}

\makebox[\textwidth]{\makebox[1.25\textwidth]{%
\begin{minipage}{0.65\textwidth}
\begin{table}[H]
 \begin{tabular}{|l|l|l|l|l|l|l|}
\hline
\multirow{2}{*}{Query} &
  \multicolumn{3}{c|}{Prediction} &
  \multicolumn{3}{c|}{Ground truth} \\
 & I & T & N & I & T & N \\
\hline
shops location & 0.98 & 0.00 & 0.02 & 0.00 & 0.00 & 1.00 \\
\hline
causes hemochromatosis & 1.00 & 0.00 & 0.00 & 1.00 & 0.00 & 0.00 \\
\hline
business mathematics & 0.99 & 0.01 & 0.00 & 0.00 & 1.00 & 0.00 \\
\hline
market boots & 0.00 & 0.47 & 0.53 & 0.00 & 1.00 & 0.00 \\
\hline
avenel hats & 0.01 & 0.99 & 0.00 & 0.00 & 0.00 & 1.00 \\
\hline
footpath maps & 0.00 & 0.00 & 1.00 & 0.00 & 0.00 & 1.00 \\
\hline
gps location & 0.61 & 0.00 & 0.39 & 0.00 & 0.00 & 1.00 \\
\hline
flute price & 0.00 & 1.00 & 0.00 & 0.00 & 1.00 & 0.00 \\
\hline
indonesien bilder & 0.95 & 0.05 & 0.01 & 0.00 & 0.00 & 1.00 \\
\hline
padlock reviews & 1.00 & 0.00 & 0.00 & 1.00 & 0.00 & 0.00 \\
\hline
pavers rockhampton & 0.63 & 0.20 & 0.17 & 0.00 & 0.00 & 1.00 \\
\hline
daylesford celebrant & 0.59 & 0.35 & 0.06 & 0.00 & 0.00 & 1.00 \\
\hline
john price & 0.00 & 1.00 & 0.00 & 0.00 & 0.00 & 1.00 \\
\hline
how to & 0.54 & 0.40 & 0.06 & 1.00 & 0.00 & 0.00 \\
\hline
minecraft keyrings & 0.58 & 0.21 & 0.21 & 0.00 & 0.00 & 1.00 \\
\hline
electric shop & 0.00 & 0.00 & 1.00 & 0.00 & 0.00 & 1.00 \\
\hline
edwards coaches & 0.34 & 0.64 & 0.02 & 0.00 & 0.00 & 1.00 \\
\hline
yamaha microphones & 0.95 & 0.00 & 0.05 & 0.00 & 0.00 & 1.00 \\
\hline
skechers adelaide & 0.29 & 0.11 & 0.60 & 0.00 & 0.00 & 1.00 \\
\hline
business economic & 1.00 & 0.00 & 0.00 & 0.00 & 0.00 & 1.00 \\
\hline
warwick information centre & 0.38 & 0.00 & 0.62 & 0.50 & 0.00 & 0.50 \\
\hline
reviews on hydroxatone & 1.00 & 0.00 & 0.00 & 1.00 & 0.00 & 0.00 \\
\hline
bus time t & 0.00 & 0.00 & 1.00 & 0.00 & 0.00 & 1.00 \\
\hline
moneygram near me & 0.00 & 0.00 & 1.00 & 0.00 & 0.00 & 1.00 \\
\hline
sydney opera events & 0.03 & 0.00 & 0.97 & 0.00 & 0.00 & 1.00 \\
\hline
bas statements online & 0.02 & 0.98 & 0.00 & 0.00 & 1.00 & 0.00 \\
\hline
herald newspaper online & 0.00 & 1.00 & 0.00 & 0.00 & 0.00 & 1.00 \\
\hline
house prices tamworth & 0.00 & 1.00 & 0.00 & 0.00 & 0.50 & 0.50 \\
\hline
victoria regions map & 0.00 & 0.00 & 1.00 & 0.00 & 0.00 & 1.00 \\
\hline
information about police & 1.00 & 0.00 & 0.00 & 1.00 & 0.00 & 0.00 \\
\hline
aged pension rates & 0.00 & 1.00 & 0.00 & 1.00 & 0.00 & 0.00 \\
\hline
sale watches australia & 0.00 & 1.00 & 0.00 & 0.00 & 1.00 & 0.00 \\
\hline
adelaide removalists reviews & 0.99 & 0.00 & 0.01 & 0.50 & 0.00 & 0.50 \\
\hline
kelvinator fridge reviews & 1.00 & 0.00 & 0.00 & 0.50 & 0.00 & 0.50 \\
\hline
watch perpetual calendar & 0.00 & 1.00 & 0.00 & 0.00 & 1.00 & 0.00 \\
\hline
snopes fact check & 0.00 & 1.00 & 0.00 & 0.00 & 1.00 & 0.00 \\
\hline
townsville house rentals & 0.00 & 1.00 & 0.00 & 0.00 & 0.00 & 1.00 \\
\hline
harley davidson gps & 0.00 & 0.00 & 1.00 & 0.00 & 0.00 & 1.00 \\
\hline
masters indigenous studies & 1.00 & 0.00 & 0.00 & 1.00 & 0.00 & 0.00 \\
\hline
public weighbridge perth & 0.25 & 0.00 & 0.75 & 0.00 & 0.00 & 1.00 \\
\hline
country master watch & 0.00 & 0.90 & 0.10 & 0.00 & 0.00 & 1.00 \\
\hline
ichthys project location & 1.00 & 0.00 & 0.00 & 0.00 & 0.00 & 1.00 \\
\hline
\end{tabular}
\end{table}
\end{minipage}
\begin{minipage}{0.65\textwidth}
\begin{table}[H]
 \begin{tabular}{|l|l|l|l|l|l|l|}
\hline
\multirow{2}{*}{Query} &
  \multicolumn{3}{c|}{Prediction} &
  \multicolumn{3}{c|}{Ground truth} \\
 & I & T & N & I & T & N \\
\hline
spacex stock price & 0.00 & 1.00 & 0.00 & 0.00 & 1.00 & 0.00 \\
\hline
local news penrith & 0.01 & 0.00 & 0.99 & 0.00 & 0.00 & 1.00 \\
\hline
suzuki cars review & 1.00 & 0.00 & 0.00 & 1.00 & 0.00 & 0.00 \\
\hline
dacia sandero review & 1.00 & 0.00 & 0.00 & 1.00 & 0.00 & 0.00 \\
\hline
haier fridge review & 1.00 & 0.00 & 0.00 & 1.00 & 0.00 & 0.00 \\
\hline
world cruise prices & 0.00 & 0.45 & 0.55 & 0.00 & 1.00 & 0.00 \\
\hline
ufo area 51 & 0.01 & 0.00 & 0.99 & 0.00 & 0.00 & 1.00 \\
\hline
wedding ceremony places & 0.47 & 0.09 & 0.44 & 0.00 & 0.00 & 1.00 \\
\hline
vtu mutual bank & 0.00 & 0.00 & 1.00 & 0.00 & 0.00 & 1.00 \\
\hline
water cartage prices & 0.00 & 1.00 & 0.00 & 0.00 & 1.00 & 0.00 \\
\hline
french country nighties & 0.22 & 0.01 & 0.77 & 0.00 & 0.00 & 1.00 \\
\hline
sizzler prices maroochydore & 0.01 & 0.99 & 0.00 & 0.00 & 0.50 & 0.50 \\
\hline
crystal serenity review & 1.00 & 0.00 & 0.00 & 1.00 & 0.00 & 0.00 \\
\hline
schools moonee ponds & 0.96 & 0.01 & 0.03 & 0.00 & 0.00 & 1.00 \\
\hline
quails for sale & 0.00 & 1.00 & 0.00 & 0.00 & 1.00 & 0.00 \\
\hline
math terms definitions & 0.55 & 0.45 & 0.00 & 1.00 & 0.00 & 0.00 \\
\hline
food safety accreditation & 1.00 & 0.00 & 0.00 & 0.00 & 1.00 & 0.00 \\
\hline
savers perfume prices & 0.00 & 1.00 & 0.00 & 0.00 & 1.00 & 0.00 \\
\hline
hire campervan melbourne & 0.00 & 1.00 & 0.00 & 0.00 & 0.50 & 0.50 \\
\hline
annaburg porzellan shop & 0.00 & 0.00 & 1.00 & 0.00 & 0.00 & 1.00 \\
\hline
aquos tv sharp & 0.00 & 1.00 & 0.00 & 0.00 & 0.00 & 1.00 \\
\hline
dimmeys stores queensland & 0.00 & 0.00 & 1.00 & 0.00 & 0.00 & 1.00 \\
\hline
small business questions & 0.96 & 0.00 & 0.04 & 0.00 & 1.00 & 0.00 \\
\hline
choice bed reviews & 1.00 & 0.00 & 0.00 & 1.00 & 0.00 & 0.00 \\
\hline
zenni optical australia & 0.65 & 0.34 & 0.01 & 0.00 & 0.00 & 1.00 \\
\hline
shelta umbrella online & 0.00 & 1.00 & 0.00 & 0.00 & 1.00 & 0.00 \\
\hline
king orchid care & 1.00 & 0.00 & 0.00 & 0.00 & 1.00 & 0.00 \\
\hline
tv recorder reviews & 0.60 & 0.40 & 0.00 & 1.00 & 0.00 & 0.00 \\
\hline
mondaine watch australia & 0.01 & 0.99 & 0.00 & 0.00 & 0.00 & 1.00 \\
\hline
euro rates today & 0.00 & 1.00 & 0.00 & 1.00 & 0.00 & 0.00 \\
\hline
learn vietnamese language & 0.00 & 1.00 & 0.00 & 0.00 & 1.00 & 0.00 \\
\hline
burwood property prices & 0.01 & 0.98 & 0.00 & 0.00 & 0.50 & 0.50 \\
\hline
noosa crest reviews & 1.00 & 0.00 & 0.00 & 1.00 & 0.00 & 0.00 \\
\hline
walking sticks perth & 0.04 & 0.00 & 0.95 & 0.00 & 0.00 & 1.00 \\
\hline
crewman for sale & 0.00 & 1.00 & 0.00 & 0.00 & 1.00 & 0.00 \\
\hline
tamworth information centre & 0.35 & 0.00 & 0.65 & 0.50 & 0.00 & 0.50 \\
\hline
wax look jacket & 0.01 & 0.98 & 0.01 & 0.00 & 1.00 & 0.00 \\
\hline
tint professor review & 0.99 & 0.00 & 0.01 & 1.00 & 0.00 & 0.00 \\
\hline
dungarees australia online & 0.01 & 0.99 & 0.00 & 0.00 & 0.00 & 1.00 \\
\hline
food for chickens & 1.00 & 0.00 & 0.00 & 0.00 & 1.00 & 0.00 \\
\hline
swansea house prices & 0.00 & 1.00 & 0.00 & 0.00 & 0.50 & 0.50 \\
\hline
human resource details & 0.93 & 0.03 & 0.04 & 1.00 & 0.00 & 0.00 \\
\hline
\end{tabular}
\end{table}
\end{minipage}}}

\end{appendices}

\end{document}